\pdfoutput=1
\PassOptionsToPackage{table}{xcolor}
\documentclass{article}

\usepackage[T1]{fontenc}
\usepackage{iclr2027_conference,times}

\usepackage{amsmath,amsfonts,bm}

\def\eqref#1{equation~\ref{#1}}

\def\1{\bm{1}}

\DeclareMathAlphabet{\mathsfit}{\encodingdefault}{\sfdefault}{m}{sl}
\SetMathAlphabet{\mathsfit}{bold}{\encodingdefault}{\sfdefault}{bx}{n}

\usepackage{amsmath}
\usepackage{amssymb}
\usepackage{xcolor}
\usepackage{booktabs}
\usepackage{float}
\usepackage{placeins}
\usepackage{needspace}
\usepackage{graphicx}
\usepackage{multirow}
\usepackage{url}
\usepackage{hyperref}
\usepackage{xspace}
\hypersetup{hidelinks,
  pdftitle={SEPAL: Separated Expert Pairs with Answer-Level Fusion for Reliable LLM Collaboration},
  pdfauthor={Weijie Ren, Yanwen Zhang, Hao Li, Zhuolin Qi, Hengyi Zhang, Naibo Wang}}

\definecolor{TableHeader}{HTML}{F1F4F8}
\definecolor{TableMacro}{HTML}{F7F9FC}
\definecolor{HighlightBlue}{HTML}{EDF4FC}
\definecolor{HighlightMint}{HTML}{E7F6EF}
\definecolor{HighlightRose}{HTML}{FBECEC}
\definecolor{PositiveText}{HTML}{236B52}
\definecolor{NegativeText}{HTML}{A24343}
\newcommand{\ourscell}[1]{\cellcolor{HighlightBlue}#1}
\newcommand{\bestcell}[1]{\cellcolor{HighlightBlue}\textcolor{black}{\textbf{#1}}}
\newcommand{\oraclecell}[1]{\cellcolor{HighlightMint}\textcolor{black}{\textbf{#1}}}
\newcommand{\gaincell}[1]{\cellcolor{HighlightMint}\textcolor{PositiveText}{\textbf{#1}}}
\newcommand{\losscell}[1]{\cellcolor{HighlightRose}\textcolor{NegativeText}{#1}}

\title{SEPAL: Separated Expert Pairs\\
with Answer-Level Fusion\\
for Reliable LLM Collaboration}

\author{Weijie Ren$^{1,*}$\enspace Yanwen Zhang$^{2,*}$\enspace Hao Li$^{3,*}$\enspace
Zhuolin Qi$^{1}$\enspace Hengyi Zhang$^{1}$\enspace Naibo Wang$^{1,\dagger}$\\
$^{1}$Zhejiang University\quad
$^{2}$University of Electronic Science and Technology of China\\
$^{3}$University of Science and Technology of China\quad
{\small$^{*}$Equal contribution.\quad $^{\dagger}$Corresponding author.}\\
{\small\texttt{3200101501@zju.edu.cn}, \texttt{2023091601016@std.uestc.edu.cn},}\\
{\small\texttt{haoli2101@mail.ustc.edu.cn}, \texttt{qizhuolin666@gmail.com},}\\
{\small\texttt{22651274@zju.edu.cn}, \texttt{wangnaibo@zju.edu.cn}}}

\iclrfinalcopy

\newcommand{\method}{SEPAL\xspace}

\begin{document}
\raggedbottom
\widowpenalty=10000
\clubpenalty=10000
\displaywidowpenalty=10000

\maketitle
\lhead{Preprint}

\begin{abstract}
Multi-agent collaboration lets large language models (LLMs) improve question
answering through deliberation and feedback. Yet shared discussion couples
correction with exposure to the same mistakes, which can erode the diversity
needed for voting. Self-consistency offers sampling diversity without feedback,
while single-pair Actor--Critic collaboration refines only one candidate. We
introduce \method, which assigns three private Actor--Critic teams to direct
reasoning, evidence grounding, and verification. Role-specific training gives
the teams different reasoning objectives beyond sampling variation. Each Critic
guides revisions within its own team, preventing feedback from carrying errors
across candidates. Once revision ends, majority voting combines only the final
answers, keeping the reasoning histories separate until the decision. Across
five open-weight backbones and five question-answering benchmarks, \method
improves mean accuracy by 1.81 percentage points over a matched single
Actor--Critic pair, with improvements across all five backbones. Code is
available at \url{https://github.com/zhansan114514/SEPAL}.
\end{abstract}

\begin{figure}[H]
  \centering
  \includegraphics[width=\linewidth]{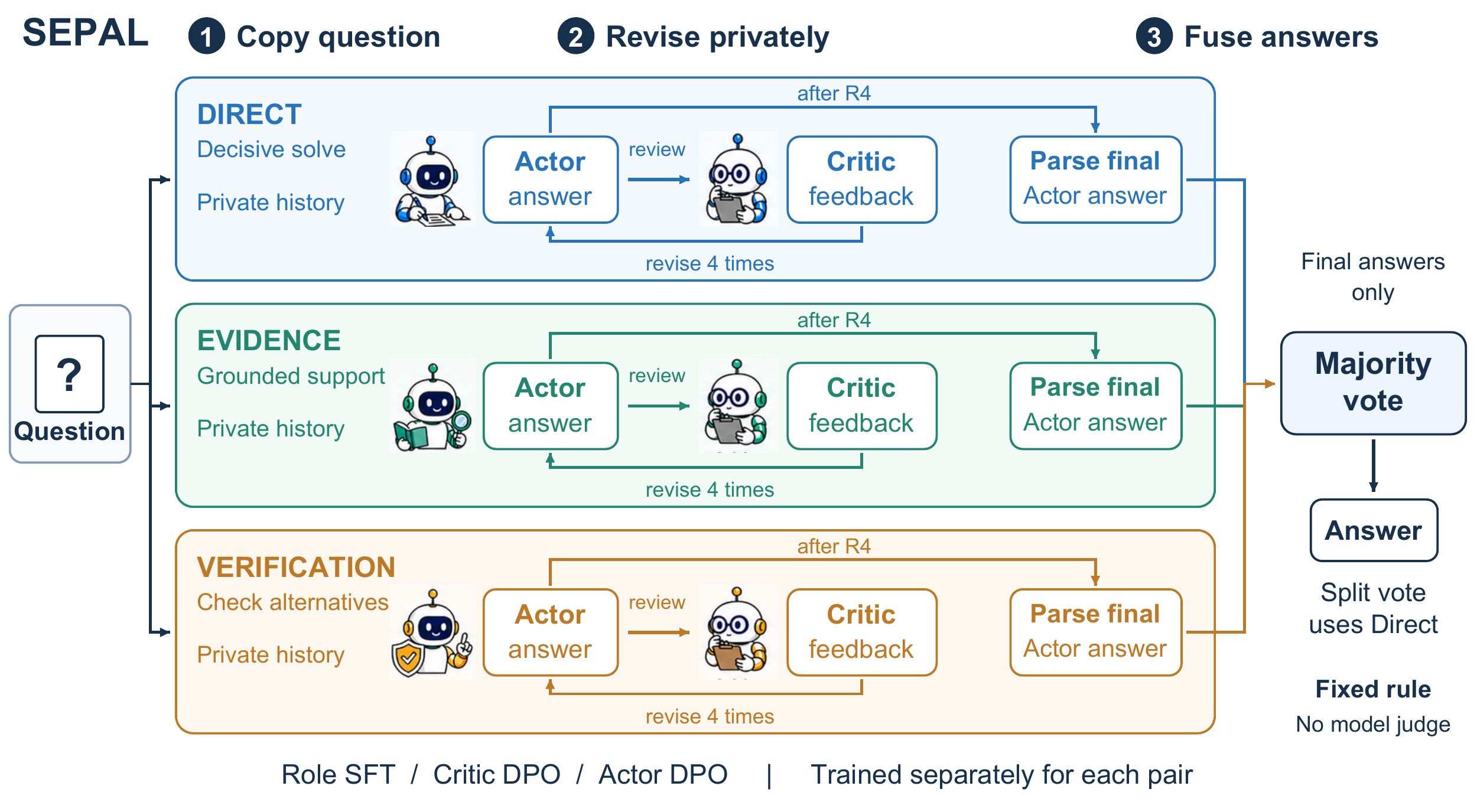}
  \caption{\textbf{SEPAL in one view.} A question enters three isolated role
  pairs, each refined through four private Critic-guided revisions (R1--R4). Only final
  parsed answers cross team boundaries. A fixed majority vote returns the
  prediction and uses Direct for split decisions.}
  \label{fig:method}
\end{figure}

\section{Introduction}

Reasoning with large language models (LLMs) increasingly relies on generating,
inspecting, and comparing intermediate solutions. Chain-of-thought prompting
makes a reasoning path explicit \citep{wei2022cot}, and self-consistency improves
reliability by sampling several paths \citep{wang2023selfconsistency}.
Multi-agent discussion adds opportunities to question an argument and revise
an answer \citep{du2024debate,liang2024mad}. For question answering, these
approaches offer both feedback that can repair mistakes and alternative
answers for the final decision.

However, existing approaches leave a gap between these two benefits. In
shared-history debate, each agent sees peer arguments, so an early error can
enter several later candidates. Self-consistency keeps sampled paths separate
but provides no corrective feedback. ACC-Collab learns an Actor and a Critic
using the correctness of subsequent Actor responses
\citep{estornell2025acccollab}, but its interaction refines a single answer.
It supplies learned feedback without a set of separately revised candidates
for voting.

The key issue is that a vote depends on how candidates fail together, as well
as how accurate they are individually. A critique can repair an answer, yet a
shared critique can also steer several agents toward the same faulty premise.
When revision increases agreement without adding independent evidence, better
individual answers can coexist with less benefit from voting. Useful
collaboration therefore depends on both individual accuracy and the errors
candidates share after feedback.

We hypothesize that restricting feedback to each team and combining only final
answers can retain revision gains while preserving useful differences between
candidates. Private feedback removes the path by which one team's faulty
argument enters another team's reasoning before the vote. The final decision
can then draw on solutions developed in separate conversation histories.

We instantiate this hypothesis in \method, \textbf{S}eparated \textbf{E}xpert
\textbf{P}airs with \textbf{A}nswer-\textbf{L}evel fusion
(Figure~\ref{fig:method}). Each of three teams contains an Actor that proposes
an answer and a Critic that guides revision. To introduce structured diversity
beyond repeated sampling of one pair, we assign different reasoning objectives.
Direct derives an answer, Evidence grounds it in relevant facts or passages,
and Verification re-solves the question and checks alternatives. These
objectives guide both answer generation and critique. Teams revise privately
before a fixed majority vote combines their final answers. Direct supplies the
fallback for split decisions, with no additional judge.

The teams share a backbone and training questions but learn separate adapters.
Role-specific supervised training initializes the Actors, followed by Critic
and Actor preference learning. A Critic's feedback is preferred during
training when it leads to more correct Actor revisions. We train each role on
its own trajectories using ACC-Collab's preference rule.

We evaluate five open-weight backbones ranging from 2B to 8B parameters on five
question-answering benchmarks. MMLU supplies all training data; the other four
benchmarks test transfer. Relative to a matched single Actor--Critic pair,
\method improves macro accuracy for every backbone by 1.06--2.22 percentage
points, averaging 1.81 points.

Our contributions are threefold.
\begin{itemize}
  \item A collaboration method that separates local Actor--Critic revision
  from final aggregation and trains three teams with distinct reasoning objectives.
  \item Evidence that revision gives the largest component gain, while final
  voting exceeds the strongest individual role in 21 of 25 evaluated settings.
  \item An analysis of diminishing returns from further revision, showing
  how later repairs are increasingly offset by regressions in correct answers.
\end{itemize}

\section{Related Work}

\paragraph{Independent sampling and late aggregation.}
Self-consistency improves chain-of-thought reasoning by sampling independent
paths and aggregating their answers \citep{wang2023selfconsistency}. Increasing
the number of agents extends the same intuition to replicated LLM calls
\citep{li2024moreagents}, while Multiagent Finetuning trains separate agents
to preserve distinct reasoning behaviors \citep{subramaniam2025multiagent}.
Tree of Thoughts preserves several partial solutions and alternates expansion
with model-based evaluation, making diversity an explicit search resource
\citep{yao2023tot}. These methods differ in where alternatives are reduced:
search may prune partial states, whereas late-voting systems keep complete
candidates until the decision interface.
Independent sampling keeps candidate histories separate, while tree search
uses intermediate evaluations to guide expansion. \method gives each complete
candidate its own learned reviewer and aggregates the revised answers.

\paragraph{Inference-time communication.}
Multi-agent debate exposes agents to peer arguments and can improve factuality
or reasoning \citep{du2024debate,liang2024mad}; ChatEval applies debate to LLM
evaluation \citep{chan2024chateval}. ReConcile aggregates
diverse LLMs through a round-table protocol \citep{chen2024reconcile}, and
Mixture-of-Agents passes outputs through layered aggregators
\citep{wang2025moa}. \method permits revision within each team and applies
a fixed answer vote after all teams finish. Its private feedback paths and
final decision rule specify the information available to each agent.

Role prompts provide a second source of diversity. CoMM assigns distinct roles
and reasoning paths and finds that independently prompted experts are important
for science reasoning \citep{chen2024comm}. Unlike its collaborative discussion,
however, \method prevents a role from seeing peer content before aggregation.
Role labels alone do not establish complementarity when a shared transcript
can synchronize the candidates.

\paragraph{Candidate selection and corrective feedback.}
LLM-Blender learns a pairwise ranker and a generative fuser to select and merge
outputs from heterogeneous models \citep{jiang2023llmblender}. That approach can
exploit information beyond exact answers, but introduces a learned selection
layer whose errors and training distribution become part of the system.
\method uses a fixed answer parser and majority rule to expose the quality
of the candidate trajectories at the decision interface.
Correction methods expose a related boundary. CRITIC grounds revision in
tool-interactive feedback \citep{gou2024critic}, while intrinsic self-correction
without external feedback can preserve or amplify reasoning errors
\citep{huang2024noselfcorrect}. Our Critics receive no tools or gold labels at
test time, but are trained using the downstream correctness of Actor
continuations. The component and round analyses therefore test whether this
learned, role-local signal repairs answers rather than assuming that another
revision prompt is beneficial.

\paragraph{Learning to revise and collaborate.}
STaR bootstraps reasoning traces from successful solutions
\citep{zelikman2022star}; Self-Refine and Reflexion use generated feedback or
verbal memory to improve later behavior \citep{madaan2023selfrefine,shinn2023reflexion}.
DPO provides a direct objective for preference learning without an explicit
reward model \citep{rafailov2023dpo}. ACC-Collab uses continuation correctness
to construct Actor and Critic preferences \citep{estornell2025acccollab}, while
Multiagent Finetuning independently specializes agents to preserve diverse
reasoning chains \citep{subramaniam2025multiagent}. We retain ACC-Collab's
continuation-valued training rule, replicate it independently for each role,
and study the resulting components rather than assuming that the full training
stack is uniformly beneficial.

\section{Method}
\label{sec:method}

We study a question $x$ with gold answer $y$. A collaborative system produces
candidate trajectories $\tau_i$ ending in responses $a_i$, and a deterministic
task-aware extractor maps each response to $z_i=g(a_i)$. The method is designed
for \emph{local repair}, so feedback can change a candidate rather than merely
score it; \emph{alternative preservation}, so candidate $i$ never conditions on
candidate $j$ before aggregation; and \emph{transparent fusion}, so the final
decision does not hide another generative Judge. Shared dialogue can make
several votes descendants of one error. Our goal is therefore \emph{useful
complementarity}: improve each trajectory while retaining enough residual
variation for late fusion to matter.

\subsection{Isolated Role Teams}

For role
$i\in\{D,E,V\}$, Actor $A_i$ and Critic $C_i$ form a separate team. At round
$0$, $A_i$ produces $a_i^0$ and $C_i$ returns feedback $c_i^0$. At rounds
$t=1,\ldots,4$, the Actor revises from its previous answer and feedback, then
the Critic reviews it unless $t=4$. The reported team answer is
$z_i=g(a_i^4)$, with the following local generation dependencies.

\begin{equation}
 a_i^0\sim A_i(\cdot\mid x),\qquad
 c_i^t\sim C_i(\cdot\mid x,a_i^t),\qquad
 a_i^{t+1}\sim A_i(\cdot\mid x,a_i^t,c_i^t).
 \label{eq:isolation}
\end{equation}

Thus a team never receives another team's response, rationale, confidence, or
adapter. Direct emphasizes a decisive derivation; Evidence grounds its answer
in the relevant definition, fact, or passage; Verification independently
re-solves the question and checks alternatives. The exact instructions are in
Appendix~\ref{app:prompts}.

Equation~\ref{eq:isolation} defines the implemented prompt dependencies.
Role prefixes are included in $A_i$ and $C_i$, and $t=0,\ldots,3$ for the
revision transition. The three pairs share pretrained weights and training
questions, so their errors can remain correlated. Separate histories remove
cross-pair text from generation; Section~\ref{sec:diagnostics} measures the
agreement that remains at the final decision.

\subsection{Why Three Encapsulated Pairs?}

Three is the smallest odd ensemble that supports a strict majority and a
nontrivial diversity analysis. The roles vary the route to an answer while
receiving the same full question and base prompt. This gives a fixed
inference budget of three five-round pair trajectories followed by a vote.

\begin{figure}[!t]
  \centering
  \includegraphics[width=\linewidth]{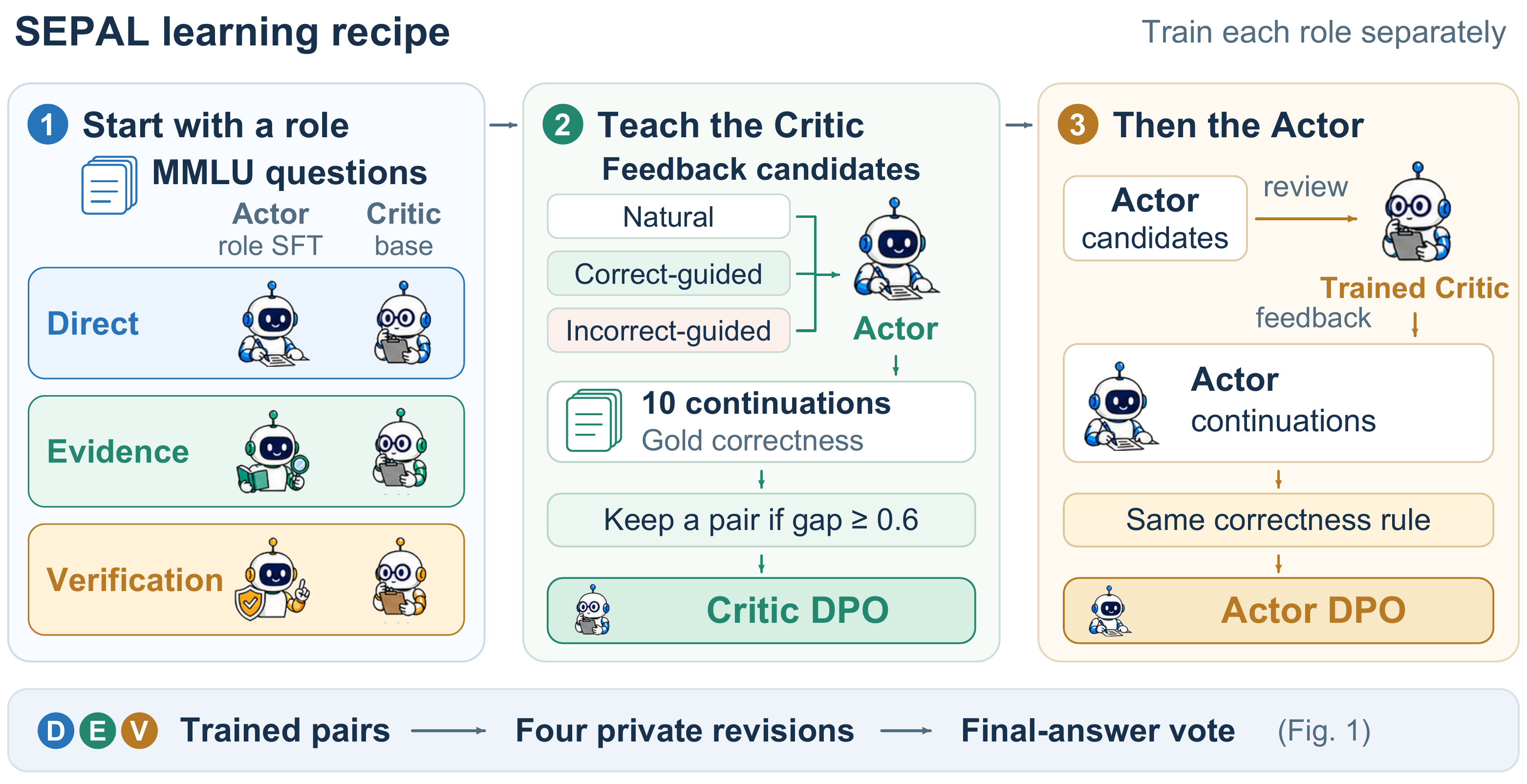}
  \caption{\textbf{Training the private teams.} Role SFT initializes Actors;
  Critics start from the base model. Feedback is valued by Actor continuation
  correctness. Critic DPO precedes Actor preference construction and Actor DPO.
  Each role follows this sequence separately before inference in Figure~\ref{fig:method}.}
  \label{fig:training}
\end{figure}

\subsection{Balanced Role Initialization}

Training begins with role-specific Actor SFT (Figure~\ref{fig:training}). For each backbone, the
base model generates candidates for 10,000 MMLU auxiliary-training questions
under every role at temperatures $0.4$, $0.7$, and $1.0$. A target is eligible
only when its extracted answer is correct and its generation is not truncated.
We keep at most one target per question--role pair and intersect question IDs
across all three roles. Consequently, roles within a backbone receive the same
questions, target count, and number of SFT updates. The retained count per role
is 7,847 for Llama, 8,313 for Qwen2.5, 6,902 for Gemma, 7,755 for Phi,
and 7,005 for Mistral. Critics start from the unadapted backbone.

\subsection{Continuation-Valued Preference Learning}

Every team processes all 1,531 MMLU validation questions, using five sampled
trajectories per question for Mistral and one for each other backbone.
It follows the ACC-Collab training order: construct Critic preferences, train the
Critic, construct Actor preferences with that Critic, then train the Actor
\citep{estornell2025acccollab}. For an Actor state $(x,a)$, the generator
samples natural feedback $c^0$, feedback guided toward a correct answer $c^+$,
and feedback guided toward an incorrect answer $c^-$. Candidate feedback is
valued through \mbox{$K=10$ Actor continuations:}

\begin{equation}
 \widehat{R}(c\mid x,a)=\frac{1}{K}\sum_{k=1}^{K}
 \mathbf{1}\!\left[g(a'_k)=y\right],\qquad
 a'_k\sim A_i(\cdot\mid x,a,c).
 \label{eq:continuation}
\end{equation}

The ordered rule retains $(c^+,c^0)$ when its reward gap is at least
$\epsilon=0.6$; otherwise it retains $(c^0,c^-)$ when that gap is at least
$\epsilon$. Each state therefore contributes at most one Critic comparison,
and weak contrasts are discarded. After Critic DPO, Actor candidates are valued
by sampling natural Critic feedback followed by an Actor continuation. The
Actor stage uses the same margin rule. The score ties feedback quality to
the Actor's next answer. Gold labels supply training supervision; inference
follows the private path in Figure~\ref{fig:method}.

For either stage, a retained preferred/dispreferred pair $(u^+,u^-)$ is trained
with DPO \citep{rafailov2023dpo}:

\begin{align}
 \mathcal{L}_{\mathrm{DPO}}(\theta)=
 -\mathbb{E}\log\sigma\!\left(\beta\left[
 \log\frac{\pi_\theta(u^+\mid s)}{\pi_{\mathrm{ref}}(u^+\mid s)}-
 \log\frac{\pi_\theta(u^-\mid s)}{\pi_{\mathrm{ref}}(u^-\mid s)}
 \right]\right)+\lambda\mathcal{L}_{\mathrm{NLL}},
 \label{eq:dpo}
\end{align}

with $\beta=0.1$ and $\lambda=1$. Here $s$ is the stage-specific prompt,
$\pi_{\mathrm{ref}}$ is the fixed reference policy, and
$\mathcal{L}_{\mathrm{NLL}}$ is the mean token negative log-likelihood of the
preferred completion $u^+$. Actor DPO starts from the role-SFT adapter. Source questions
are not partitioned among roles: every team receives the complete source set
but produces independent trajectories and a data-dependent number of retained
pairs.

\subsection{Judge-Free Late Fusion}

After round 4, the system counts valid answers. If an answer receives at least
two votes, it is returned; otherwise the Direct answer is the fixed fallback:

\begin{equation}
 \widehat{y}=\begin{cases}
 m, & |\{i:z_i=m\}|\geq2,\\
 z_D, & \text{otherwise}.
 \end{cases}
 \label{eq:vote}
\end{equation}

The fallback is fixed before evaluation. Critic text is never a vote, and no
model is invoked to adjudicate disagreements. Compared with one ACC-Collab
team, \method uses three pair trajectories; the vote adds negligible cost.

\section{Experiments}
\label{sec:experiments}

\subsection{Experimental Setup}

\paragraph{Models, data, and benchmarks.}
We evaluate Meta-Llama-3-8B-Instruct \citep{grattafiori2024llama3},
Qwen2.5-3B-Instruct \citep{yang2024qwen25}, Gemma-2-2B-it
\citep{gemmateam2024gemma2}, Phi-4-mini-instruct
\citep{microsoft2025phi4mini}, and Mistral-7B-Instruct-v0.3
\citep{jiang2023mistral7b}. MMLU contains 57 academic subjects
\citep{hendrycks2021mmlu}. BoolQ is yes/no reading comprehension
\citep{clark2019boolq}; BBH collects challenging BIG-Bench tasks
\citep{suzgun2023bbh}; SciQ is multiple-choice science QA
\citep{welbl2017sciq}; and ARC combines the official Easy and Challenge test
splits \citep{clark2018arc}. We use the available examples for four datasets
and a category-stratified 1,260-example BBH subset; only MMLU supplies
optimization data.

\begin{table}[h]
  \centering
  \small
  \setlength{\tabcolsep}{4.0pt}
  \begin{tabular}{llrl}
\toprule
\rowcolor{TableHeader}
Dataset & Split & $N$ & Use \\
\midrule
MMLU & test & 14,042 & in-domain \\
BoolQ & validation & 3,270 & transfer \\
BBH & 22-category stratified & 1,260 & transfer \\
SciQ & test & 1,000 & transfer \\
ARC & Easy + Challenge test & 3,548 & transfer \\
\bottomrule
\end{tabular}

  \caption{Evaluation datasets. Transfer rows supply no training examples.}
  \label{tab:data}
\end{table}

\paragraph{Baselines and metric.}
Direct is one response from the unadapted instruction model. Debate uses an
untrained Actor and Critic for the same five-round protocol. SoM-2 and SoM-4
use two or four symmetric untrained agents that observe peer responses; the
metric is mean final-round accuracy over agents, following their original
evaluation protocol.
ACC is a single trained Actor--Critic pair using the same MMLU source data,
parser, splits, and hyperparameters as \method. The primary metric is exact
match after task-aware normalization, and macro accuracy is the unweighted mean
over five datasets. All comparisons are descriptive, using one fixed
realization per model--method--dataset cell under the recorded protocol.

\begin{table}[!t]
  \centering
  \normalsize
  \setlength{\tabcolsep}{4.0pt}
  \begin{tabular}{llrrrrrr}
\toprule
\rowcolor{TableHeader}
Model & Dataset & Direct & Debate & SoM-2 & SoM-4 & ACC & \method \\
\midrule
\multirow{6}{*}{Llama-3-8B} & BoolQ & 77.34 & 76.76 & \textbf{78.98} & 78.74 & 76.54 & \ourscell{76.70} \\
 & MMLU & 62.41 & 63.55 & 63.39 & 63.33 & 65.00 & \bestcell{66.93} \\
 & BBH & 49.52 & 50.00 & 50.52 & 51.31 & 53.17 & \bestcell{56.67} \\
 & SciQ & 91.90 & 92.00 & 92.45 & 92.03 & 91.50 & \bestcell{93.40} \\
 & ARC & 88.30 & 88.92 & 89.04 & 88.65 & 89.04 & \bestcell{90.78} \\
\rowcolor{TableMacro}
 & Macro & 73.89 & 74.25 & 74.87 & 74.81 & 75.05 & \bestcell{76.90} \\
\midrule
\multirow{6}{*}{Qwen2.5-3B} & BoolQ & 65.17 & 69.30 & 67.58 & 68.21 & 73.30 & \bestcell{77.71} \\
 & MMLU & 65.80 & 65.67 & 65.81 & 66.04 & 67.45 & \bestcell{68.37} \\
 & BBH & 50.87 & 49.60 & 53.45 & 54.86 & 51.98 & \bestcell{55.63} \\
 & SciQ & \textbf{92.70} & 92.30 & 92.05 & 91.77 & 91.80 & \ourscell{91.90} \\
 & ARC & 89.49 & 90.90 & 89.56 & 90.17 & 90.84 & \bestcell{92.42} \\
\rowcolor{TableMacro}
 & Macro & 72.81 & 73.55 & 73.69 & 74.21 & 75.07 & \bestcell{77.21} \\
\midrule
\multirow{6}{*}{Gemma-2-2B} & BoolQ & 71.47 & 79.85 & \textbf{80.64} & 79.61 & 80.40 & \ourscell{80.24} \\
 & MMLU & 56.48 & 57.68 & 58.19 & 58.12 & 58.62 & \bestcell{59.50} \\
 & BBH & 41.83 & 40.16 & 38.21 & 37.48 & 42.54 & \bestcell{44.05} \\
 & SciQ & 89.20 & 90.80 & 90.65 & 90.38 & 89.90 & \bestcell{91.80} \\
 & ARC & 84.67 & 86.78 & \textbf{87.02} & 86.94 & 85.82 & \ourscell{87.01} \\
\rowcolor{TableMacro}
 & Macro & 68.73 & 71.05 & 70.94 & 70.50 & 71.46 & \bestcell{72.52} \\
\midrule
\multirow{6}{*}{Phi-4-mini} & BoolQ & 70.31 & 80.40 & 84.16 & 84.01 & 82.97 & \bestcell{85.26} \\
 & MMLU & 67.29 & 70.25 & 69.37 & 70.17 & 71.41 & \bestcell{73.04} \\
 & BBH & 52.78 & 56.03 & 55.95 & 57.40 & 57.46 & \bestcell{60.08} \\
 & SciQ & 90.70 & 92.30 & 90.05 & 91.27 & 91.80 & \bestcell{92.90} \\
 & ARC & 89.88 & 92.70 & 91.05 & 91.92 & 91.80 & \bestcell{93.21} \\
\rowcolor{TableMacro}
 & Macro & 74.19 & 78.34 & 78.12 & 78.95 & 79.09 & \bestcell{80.90} \\
\midrule
\multirow{6}{*}{Mistral-7B} & BoolQ & 75.78 & 79.20 & 79.48 & 79.12 & 80.18 & \bestcell{83.88} \\
 & MMLU & 56.95 & 59.17 & 58.10 & 57.90 & 59.69 & \bestcell{62.09} \\
 & BBH & 42.14 & 44.13 & 44.17 & 44.03 & 46.35 & \bestcell{48.10} \\
 & SciQ & 85.10 & 84.90 & 86.85 & 86.83 & 88.60 & \bestcell{89.10} \\
 & ARC & 83.77 & 84.58 & 84.72 & 85.03 & 84.24 & \bestcell{87.01} \\
\rowcolor{TableMacro}
 & Macro & 68.75 & 70.40 & 70.66 & 70.58 & 71.81 & \bestcell{74.03} \\
\bottomrule
\end{tabular}

  \caption{Accuracy (\%). ACC is the matched single-team implementation and
  \method uses three teams. Black bold marks each row's best value; pale blue
  identifies \method throughout. Macro averages the
  five datasets.}
  \label{tab:main}
\end{table}

\paragraph{Implementation.}
We use model-native chat templates, bfloat16 weights, and vLLM
\citep{kwon2023vllm} on up to four 80\,GB NVIDIA A800 GPUs. Exact generation
limits, optimizer settings, adapter configuration, seeds, prompts, and answer
handling are recorded in Appendix~\ref{app:protocol}. The adapters follow the
LoRA parameterization \citep{hu2022lora}. The matched ACC baseline
uses one Actor--Critic trajectory, whereas \method runs three role-local
trajectories before voting; other baselines keep their original call patterns.
The main comparison therefore follows each method's own protocol rather
than imposing an artificial equal-call budget.

These choices make the evaluation answer three connected questions. Does late
fusion improve on a matched learned pair? Do private trajectories retain useful
alternatives at the decision point? Which part of the training and revision
cycle produces the gain? The main table answers the first question; the
decision, ablation, and revision analyses trace each role's trajectory back to
the final vote.

\subsection{Main Results}

The first question has a consistent answer. Table~\ref{tab:main} shows that
\method exceeds matched ACC in 24 of 25 model--dataset cells. Macro accuracy
improves for all five backbones, by 1.06--2.22 points and 1.81 points on
average. Phi attains the highest absolute macro accuracy (80.90), while Mistral
has the largest gain over ACC (+2.22). Gemma BoolQ is the only negative cell
($-0.15$ points).

The improvement also transfers beyond the optimization distribution. Training
uses MMLU only, while BBH, SciQ, and ARC improve over ACC for all five
backbones; MMLU itself improves in every case. BoolQ is mixed. Lower-cost
baselines lead on four individual rows: SoM-2 on Llama BoolQ, Gemma BoolQ,
and Gemma ARC, and Direct on Qwen2.5 SciQ. These baselines use their own
recorded inference protocols, so the comparison reflects both accuracy and
the practical cost of the full recorded protocol.

The result is therefore not explained by a uniformly stronger individual role.
Some cheaper baselines remain best on individual cells, while the three-role
vote raises the macro score for every backbone. We next examine whether that
gain comes from retaining different answers until the final decision.

\subsection{Decision Analysis}
\label{sec:diagnostics}

Aggregate accuracy cannot distinguish complementarity from three copies of one
policy. We therefore measure majority coverage, unanimity, oracle-any-role
accuracy, fallback use, and per-role accuracy for every final decision record.

\begin{figure}[!htbp]
  \centering
  \includegraphics[width=\linewidth]{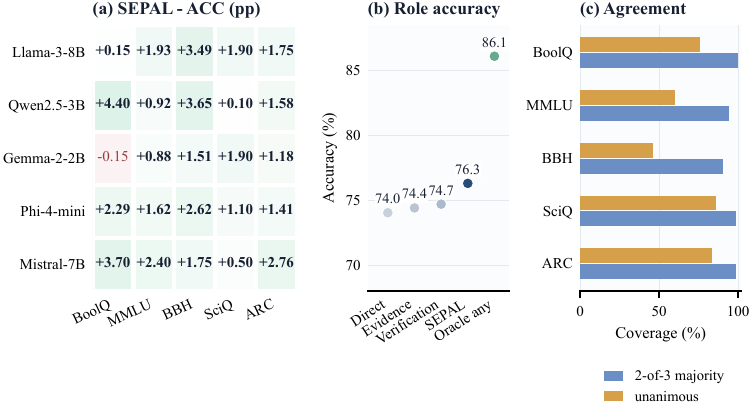}
  \caption{\textbf{Decision diagnostics across all 25 cells.}
  (a) Accuracy change from ACC to \method. (b) Mean role, vote, and
  oracle-any-role accuracy. (c) Majority coverage and unanimity by dataset,
  averaged over backbones.}
  \label{fig:decision-diagnostics}
\end{figure}

Figure~\ref{fig:decision-diagnostics} shows how these gains reach the final
decision. Direct, Evidence, and Verification average 74.05, 74.42, and
74.71\% accuracy; the vote reaches 76.31\%, exceeding the strongest role in
21 of 25 cells and by 0.62 points on average.

Consensus and unanimity are different. A two-of-three answer exists for
96.41\% of examples on average, so the Direct fallback is used only 3.59\% of
the time. Yet all three roles agree on 70.34\%. BBH is the clearest case:
majority coverage is 90.51\%, while unanimity is only 46.27\%. The system can
therefore make a stable decision while retaining substantial role-level
variation. Oracle-any-role accuracy reaches 86.12\%, exposing a 9.81-point
selection gap. This gap quantifies the opportunity for a future calibrated
router or verifier evaluated on a separate validation protocol.

Pairwise agreement covers 78.70--79.22\% of examples, with shared-answer
accuracy of 81.80--82.09\%. Thus pairs retain different answers on roughly
one fifth of questions, while agreement predicts greater reliability.
Appendix~\ref{app:diagnostics} gives the full role and decision statistics.

The vote improves on the individual roles, while the oracle gap shows that
useful answers still go unselected. We now turn to the candidates themselves
and examine which stages of training and revision make them more accurate.

\subsection{Ablations}

We first ask whether the gain comes from stronger one-shot Actors or from
interaction with Critics. Table~\ref{tab:ablation-macro} follows the same
three-role team through initialization, preference learning, and revision.
SFT-only Actors provide the starting point. SFT+Base-C and SFT+Trained-C add
private feedback from a base or preference-trained Critic, respectively.
Full-R0 evaluates the fully trained Actors before feedback, and Full-R4
includes all four revisions. No-SFT removes role initialization from the full
pipeline. The final vote and all 25 evaluation cells are shared across variants.

\begin{table}[!htbp]
  \centering
  \small
  \setlength{\tabcolsep}{3.7pt}
  \begin{tabular}{lrrrrrr}
\toprule
\rowcolor{TableHeader}
Model & SFT & Base-C & Trained-C & Full-R0 & No-SFT & Full-R4 \\
\midrule
Llama-3-8B & 71.95 & 76.35 & 76.72 & 72.27 & \textbf{77.79} & \ourscell{76.90} \\
Qwen2.5-3B & 75.57 & 76.71 & \textbf{77.79} & 75.98 & 75.98 & \ourscell{77.21} \\
Gemma-2-2B & 69.58 & 71.96 & 72.31 & 69.70 & 71.98 & \ourscell{\textbf{72.52}} \\
Phi-4-mini & 74.37 & 80.29 & \textbf{81.03} & 75.41 & 80.41 & \ourscell{80.90} \\
Mistral-7B & 70.20 & 73.38 & 73.66 & 69.60 & 73.65 & \ourscell{\textbf{74.03}} \\
\midrule
\rowcolor{TableMacro}
Mean & 72.34 & 75.74 & 76.30 & 72.59 & 75.96 & \ourscell{\textbf{76.31}} \\
\bottomrule
\end{tabular}

  \caption{Macro accuracy (\%) over five datasets. Black bold marks the best
  variant for each backbone; pale blue identifies Full-R4.}
  \label{tab:ablation-macro}
\end{table}

\begin{figure}[!t]
  \centering
  \includegraphics[width=\linewidth]{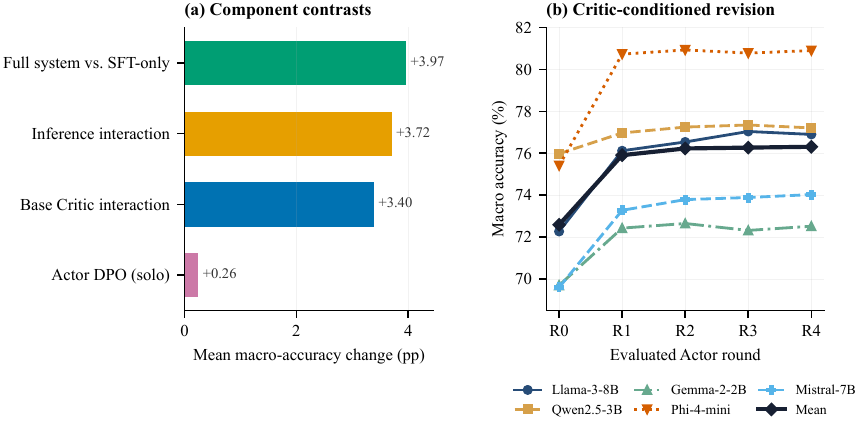}
  \caption{Component ablations and revision rounds. Left: observed macro-accuracy contrasts
  averaged over five backbones. Right: majority-vote macro accuracy by Actor
  round, with the first Critic-conditioned revision at R1.}
  \label{fig:ablation-rounds}
\end{figure}

Critic feedback produces the largest improvement
(Figure~\ref{fig:ablation-rounds}). The same trained Actors gain +3.72 macro
points from Full-R0 to Full-R4, with improvements in 23 of 25 cells. Feedback
is already useful without Critic preference learning. SFT+Base-C improves on
SFT-only by +3.40 points, also in 23 cells. By comparison, Full-R0 gains just
+0.26 points over SFT-only before any feedback, improving 17 cells. Better
initial answers account for only a small part of the full system's gain.

Preference learning has a smaller effect once feedback is available.
Replacing the base Critic with the trained Critic adds +0.57 points on
average; subsequently training the Actor adds +0.01. These averages hide
differences among backbones. SFT+Trained-C leads on Qwen2.5 and Phi, while
No-SFT leads on Llama. The complete pipeline improves on SFT-only by +3.97
points and in 24 of 25 cells, but the best configuration depends on the backbone.

These Actor comparisons address different questions. Full-R0 versus SFT-only
measures the update before any feedback. Full-R4 versus SFT+Trained-C measures
the same update when a trained Critic can respond. Reading the two together
separates initial-answer quality from the benefit that remains after feedback.

All variants keep the three private role histories and the same final vote.
The ablations therefore compare candidate generation under a fixed fusion
rule. The remaining question is how much further improvement comes from
repeating the revision step. All contrasts use unrounded aggregates;
Appendix~\ref{app:ablations} gives the full per-dataset results.

\subsection{Revision Dynamics}
\label{sec:revision}

How long should a team keep revising? The first Critic-conditioned exchange
raises mean accuracy from 72.59 to 75.91\%, capturing 89.2\% of the final
3.72-point gain. The per-example decision records separate two processes
behind this trajectory. At R1, 6.50\% of all examples change from wrong to
right, while 3.19\% change from right to wrong. By R4, these fractions reach
7.79\% and 4.07\%. Every transition is measured against the same R0 decision.

From R1 to R4, the corrected fraction rises by 1.29 percentage
points and the regressed fraction by 0.89 points, for a net gain of
0.40 points. These changes use unrounded cell averages against the same R0
baseline. A useful stopping policy must account for both effects, since
additional feedback can rescue an unresolved error or overturn an answer
that was already correct.

Later rounds keep correcting errors, but increasingly offset those repairs
with regressions. The net benefit also depends on the dataset. From R1 to R4,
BoolQ's gain over R0 grows from 5.68 to 6.91 points; BBH's falls from 4.02 to
3.33. The common R4 protocol therefore leaves room for a stopping rule
validated on held-out data. The released decisions preserve all these
transitions, including the changes that reduce accuracy.

All methods use model-native chat templates and the same answer extractor.
Appendix~\ref{app:extraction} describes parser corrections and uniform
re-scoring of the cached raw generations.

\section{Discussion and Conclusion}

The decision records separate two ways a team can fail. When all three roles
are wrong, stronger candidates or external evidence are needed. When a correct
role loses the vote, selection is the bottleneck. The 9.81-point oracle-any-role
gap identifies room for a router trained on disjoint validation data.
A selector restricted to the three final answers cannot exceed this oracle,
giving future routing experiments a measurable ceiling.
Correction and regression counts provide a second target: deciding when to
stop revising. Private histories make both interventions traceable to the
candidates that produced the final answer.

\paragraph{Limitations.}
\label{sec:limitations}

\method uses three pair trajectories and reports one fixed run per cell on
five 2B--8B backbones and short-answer QA tasks. Repeated training seeds,
comparisons at equal compute, and a direct communication-boundary ablation
would clarify robustness and the source of \mbox{the ensemble gain.}

\method keeps critique inside three private Actor--Critic teams and fuses only
their final answers. Across five backbones, it improves macro accuracy over
matched single-team ACC-Collab and wins 24 of 25 model--dataset comparisons.
Critic-conditioned revision produces the largest measured component gain,
most of it after the first exchange. Late voting then improves on the
strongest role in most evaluated cells. The resulting recipe is to revise
candidates locally and fuse completed answers. Released records trace repairs
and selection errors to individual questions.

\clearpage
\subsection*{Ethics statement}

The study uses public benchmark questions and open-weight language models; it
does not involve human subjects or newly collected personal data. The method
can nevertheless inherit factual errors, social biases, and unsafe behavior
from its backbones, and majority agreement is not a guarantee of truth. Its
additional computation also has environmental cost. The reported system should
not be used as an autonomous decision maker in high-stakes settings without
domain-specific validation and human oversight.

\subsection*{Reproducibility statement}

Code, configurations, and result records are publicly available at
\url{https://github.com/zhansan114514/SEPAL}. The repository contains the
source code with its test suite, 50 training and evaluation configurations
for the five backbones, resolved per-role configurations, aggregate metrics
for every reported cell, the result matrices as CSV files, and 125 compressed
per-example decision files. Two scripts recompute the reported numbers from
these records without model inference. Appendix~\ref{app:protocol} gives the
checkpoints, data splits, hyperparameters, and generation settings;
Appendix~\ref{app:prompts} gives the prompts; and Appendix~\ref{app:ablations}
reports every ablation cell.

\subsection*{AI use statement}

During the experiments, we used large language models (LLMs) as an auxiliary
tool for experiment monitoring and management. Specifically, the LLMs were
used to monitor experiment logs and runtime information, identify potential
execution issues or anomalies, and assist in reporting the status of ongoing
experiments. The LLMs did not determine the research questions, experimental
methodology, hyperparameter settings, or final experimental conclusions. All
experimental configurations, result verification, analysis, and scientific
conclusions were determined and validated by the authors. We take full
responsibility for the final content and results of this work.

\bibliography{references}
\bibliographystyle{iclr2027_conference}

\clearpage
\appendix

\section{Experimental Details}
\label{app:protocol}

This appendix lists the settings behind every result cell: (i) the model
checkpoint and adapter state, (ii) the dataset split and expected sample count,
(iii) the prompt, parser, and decision rule, and (iv) the resolved training or
evaluation configuration. A new run is directly comparable to our results only
when all four match the released records.

\subsection{Checkpoints and Data Lifecycle}

Table~\ref{tab:checkpoints} records the five checkpoint identities.

\begin{table}[H]
  \centering
  \small
  \setlength{\tabcolsep}{6pt}
  \begin{tabular}{ll}
    \toprule
    \rowcolor{TableHeader}
    Backbone & Checkpoint identifier \\
    \midrule
    Llama-3-8B & \nolinkurl{meta-llama/Meta-Llama-3-8B-Instruct} \\
    Qwen2.5-3B & \nolinkurl{Qwen/Qwen2.5-3B-Instruct} \\
    Gemma-2-2B & \nolinkurl{google/gemma-2-2b-it} \\
    Phi-4-mini & \nolinkurl{microsoft/Phi-4-mini-instruct} \\
    Mistral-7B & \nolinkurl{mistralai/Mistral-7B-Instruct-v0.3} \\
    \bottomrule
  \end{tabular}
  \caption{Public checkpoint identities used in every reported run.}
  \label{tab:checkpoints}
\end{table}

Deployments used offline mirrors with the same model configuration and weights.
The three phases consume disjoint dataset uses:

\begin{enumerate}
  \item \textbf{Role initialization.} For each backbone, 10,000 questions are
  sampled from MMLU \texttt{auxiliary\_train}. Candidate responses are generated
  separately under the Direct, Evidence, and Verification prefixes.
  \item \textbf{Preference construction.} All Critic and Actor comparisons use
  the 1,531-question MMLU validation split. Every role receives the complete
  source set, including every subject.
  \item \textbf{Evaluation.} MMLU uses its 14,042-item test split. BoolQ, SciQ,
  and ARC use the official splits listed in Table~\ref{tab:data}. BBH uses a
  1,260-item subset sampled at approximately 25\% from each of 22 evaluable
  categories.
\end{enumerate}

Only MMLU supplies optimization examples. BoolQ, BBH, SciQ, and ARC are
transfer evaluations. Gold evaluation labels are passed to the scorer, never
to the generation prompt, role selector, vote, or fallback. Direct is fixed
before each full-system evaluation.

\subsection{Generation and Optimization Settings}

Table~\ref{tab:hyperparameters} expands the compact implementation paragraph in
the main text. Generation uses model-native chat templates with thinking mode
disabled and bfloat16 weights. Phi uses a 4,096-token model context; the other
backbones use 8,192 tokens. The
DPO runner enforces a 4,096-token total training limit. The recorded
3,072/1,024 prompt/completion budgets are configuration metadata; the released
TRL runner applies the total limit. Both role SFT and DPO use LoRA rank 256
and \mbox{scaling factor 512.}

\begin{table}[H]
  \centering
  \small
  \setlength{\tabcolsep}{4.5pt}
  \begin{tabular}{lrrr}
\toprule
\rowcolor{TableHeader}
Setting & Role SFT & Preference / DPO & Evaluation \\
\midrule
Source split & MMLU auxiliary train & MMLU validation & benchmark-specific \\
Source questions & 10,000 & 1,531 & all configured \\
Generation temperature & $\{0.4,0.7,1.0\}$ & 0.7 & 0.7 \\
Top-$p$ & 0.9 & 0.9 & 0.9 \\
Maximum new tokens & 1,024 & 1,024 & 1,024 \\
Epochs & 1 & 3 & Not used \\
Learning rate & $5.0\!\times\!10^{-5}$ & $1.41\!\times\!10^{-5}$ & Not used \\
Effective batch size & 16 & 4 & Not used \\
Warmup ratio & 0.10 & 0.03 & Not used \\
Weight decay & 0.01 & 0.10 & Not used \\
Maximum gradient norm & 1.0 & 0.3 & Not used \\
LoRA rank / scaling & 256 / 512 & 256 / 512 & loaded adapters \\
DPO $\beta$ / NLL weight & Not used & 0.1 / 1.0 & Not used \\
Training token limit & 4,096 & 4,096 & Not used \\
\bottomrule
\end{tabular}

  \caption{Resolved settings for role initialization, preference optimization,
  and evaluation.}
  \label{tab:hyperparameters}
\end{table}

Preference construction uses $K=10$ Actor continuations per feedback candidate
and margin $\epsilon=0.6$. Mistral uses five independently sampled preference
trajectories per source question; the other backbones use one. These are
training trajectories, and every reported evaluation cell uses one trial.
DPO uses sigmoid loss, AdamW, gradient
checkpointing, and optimizer-state-aware resumption. SFT and DPO checkpoints
retain only the latest scheduled state. No evaluation score is used to select
an optimizer checkpoint.

\begin{table}[H]
  \centering
  \small
  \begin{tabular}{lrrr}
\toprule
\rowcolor{TableHeader}
Model & SFT / role & Critic pairs & Actor pairs \\
\midrule
Llama-3-8B & 7,847 & 827--2,041 & 437--828 \\
Qwen2.5-3B & 8,313 & 429--771 & 960--1,032 \\
Gemma-2-2B & 6,902 & 494--593 & 704--927 \\
Phi-4-mini & 7,755 & 276--533 & 880--975 \\
Mistral-7B & 7,005 & 1,077--2,223 & 2,811--4,813 \\
\bottomrule
\end{tabular}

  \caption{Observed optimization-data yields. Preference entries give the
  role-wise minimum--maximum under the fixed margin rule.}
  \label{tab:yields}
\end{table}

Balanced SFT retains only question IDs with an eligible target for every role
within a backbone. Consequently, the three roles receive identical SFT counts
and question identities even though their generated targets differ. Preference
yields are allowed to differ because the continuation-valued margin is applied
independently to each role state.

\subsection{Runtime, Seeds, and Execution Order}

The original environment used Python 3.10, vLLM inference, and up to four
80\,GB NVIDIA A800 GPUs. Prefix caching is enabled; eager execution is disabled;
Llama and Gemma allow 8,192 batched tokens and 128 sequences, while Qwen2.5,
Phi, and Mistral allow 32,768 batched tokens and 256 sequences. The base
seed is 42. Direct, Evidence, and Verification use role offsets 0, 10,000, and
20,000 so that they cannot accidentally share a random stream.

For every role, the order is fixed: build Critic preferences, train the Critic,
build Actor preferences using that trained Critic, then train the Actor. At
evaluation, all three role records are aligned by trial, sample index, sample
identifier, task type, and gold-label signature before voting. Any mismatch
in length, sample identity, parser version, or role configuration stops
aggregation.

\section{Prompt Templates and Role Instructions}
\label{app:prompts}

Each prompt consists of an ACC-Collab base template plus exactly one role
prefix. The released prompt identity concatenates the base version, the role
prefix version, and the role name. This prevents a prompt edit from being
silently treated as the same experiment.

\subsection{Base Actor and Critic Templates}

For a multiple-choice item, the initial Actor receives this semantic template
after its role prefix:

\begin{quote}\small
``Please answer the following multiple choice question as accurately as
possible. You must provide an extremely brief justification for your answer,
and you must give your final answer as a letter by saying `Final Answer:'.
Question: \emph{[question]}. Options: \emph{[labeled choices]}.''
\end{quote}

Each revision prompt places the previous Actor answer before its paired
Critic feedback:

\begin{quote}\small
``Several people have provided answers to a multiple choice question. Person 0
said: \emph{[previous Actor response]}. Person 1 said: \emph{[paired Critic
feedback]}. Take these answers into consideration, give an extremely brief
justification, and state the final answer as a letter.''
\end{quote}

The natural Critic template asks for brief additional details that improve the
correctness of the supplied Actor response. Guided preference candidates use
the same question and response but explicitly request details supporting a
target answer. Yes/no tasks substitute a constrained final answer of Yes or No
and include the BoolQ passage when present. These templates are identical
across roles; only the prefixes below change.

\subsection{Exact Role Prefixes}

Each following string follows the literal prefix \texttt{Role specialization}
and the role name in parentheses, then a colon. Two newline characters
separate this role instruction from the corresponding base prompt.

\paragraph{Direct Actor.}
\emph{Solve directly and concisely. Identify the decisive fact or calculation
and avoid adding speculative alternatives once the answer is supported.}

\paragraph{Direct Critic.}
\emph{Check the decisive fact or calculation in the Actor response. Supply only
brief missing details that improve correctness, following the original
\mbox{ACC-Collab critic style.}}

\paragraph{Evidence Actor.}
\emph{Ground the answer in the most relevant definition, fact, passage evidence,
or domain principle before selecting the final option.}

\paragraph{Evidence Critic.}
\emph{Check whether the Actor used the relevant evidence or principle correctly
and add only brief corrective evidence in the original \mbox{ACC-Collab critic style.}}

\paragraph{Verification Actor.}
\emph{Independently solve the problem, then verify the selected option against
the main alternatives or likely failure mode before stating the final answer.}

\paragraph{Verification Critic.}
\emph{Independently verify the Actor answer and its strongest alternative, then
provide only brief details that improve correctness in the original
\mbox{ACC-Collab critic style.}}

The instructions specialize how a pair approaches a question; they do not
grant external tools, retrieval, private reference material, different answer
choices, or different token budgets. Critic prefixes also preserve the base
instruction that feedback should be terse and corrective rather than a second
full answer presented to another role.

\section{Algorithms and Computational Structure}

\subsection{Continuation-Valued Preference Construction}

\begin{figure}[H]
\centering
\fbox{\begin{minipage}{0.94\linewidth}\small\raggedright
\textbf{Algorithm 1: one role-local preference stage.}\\[2pt]
\textbf{Input:} state $s=(x,a)$, gold answer $y$, Actor $A_i$, Critic $C_i$,
extractor $g$, rollout count $K=10$, margin $\epsilon=0.6$.\\
\textbf{1.} Generate natural feedback $c^0$, correct-guided feedback $c^+$,
and \mbox{incorrect-guided feedback $c^-$.}\\
\textbf{2.} For each $c$, sample $K$ one-step Actor continuations and compute
$\widehat{R}(c\mid s)=K^{-1}\sum_k\mathbf{1}[g(a'_k)=y]$.\\
\textbf{3.} If $\widehat{R}(c^+)-\widehat{R}(c^0)\geq\epsilon$, retain
$(c^+,c^0)$; else if $\widehat{R}(c^0)-\widehat{R}(c^-)\geq\epsilon$, retain
$(c^0,c^-)$; otherwise discard the state.\\
\textbf{4.} Train $C_i$ on retained Critic pairs. Rebuild comparisons for
Actor candidates using feedback from the trained $C_i$, then train $A_i$ with
the same ordered margin rule.\\
\textbf{Output:} one independently adapted Actor--Critic pair.
\end{minipage}}
\caption{Role-local preference construction. The ordered if/else rule permits
at most one retained comparison per state and stage.}
\label{alg:preference}
\end{figure}

The guided candidates are used only to construct preference comparisons; they
are not evaluation-time hints. Continuation correctness evaluates whether a
piece of feedback helps the Actor reach the gold answer, rather than whether
the feedback text resembles a \mbox{written reference critique.}

\subsection{Private Revision and Late Fusion}

\begin{figure}[H]
\centering
\fbox{\begin{minipage}{0.94\linewidth}\small\raggedright
\textbf{Algorithm 2: SEPAL inference.}\\[2pt]
\textbf{Input:} question $x$; pairs $(A_i,C_i)$ for
$i\in\{D,E,V\}$; extractor $g$.\\
\textbf{1.} In parallel, each Actor produces $a_i^0$ from $x$ and its role
prefix; its paired Critic returns $c_i^0$.\\
\textbf{2.} For $t=1,\ldots,4$, each Actor revises using only
$(x,a_i^{t-1},c_i^{t-1})$; for $t<4$, its paired Critic reviews \mbox{the new answer.}\\
\textbf{3.} Parse $z_i=g(a_i^4)$. Do not expose $a_i^t$, $c_i^t$, or $z_i$ to
another pair during Steps 1--3.\\
\textbf{4.} If a valid answer occurs at least twice among
$\{z_D,z_E,z_V\}$, return it. Otherwise return \mbox{the parsed Direct answer.}\\
\textbf{Output:} final answer $\widehat{y}$ and its decision source
(unanimous, majority, or \mbox{fixed Direct fallback).}
\end{minipage}}
\caption{Inference contains no learned Judge and no cross-pair message.}
\label{alg:inference}
\end{figure}

With five Actor rounds and one Critic response after each of the first four
Actor responses, one ACC pair uses five Actor and four Critic generations.
\method uses three such trajectories: 15 Actor and 12 Critic generations,
parallelizable across roles.
Answer extraction and voting are deterministic string operations. Direct uses
one Actor generation; SoM and Debate follow their original protocols and are
not normalized to the same number of generated tokens.

\section{Complete Main Comparisons}

Table~\ref{tab:main-deltas} expands the headline comparison into the exact
sample count, ACC accuracy, \method accuracy, and difference for every cell.
The better accuracy in each row is black bold; positive differences appear in
mint and the single negative cell in pale red.

\begin{table}[H]
  \centering
  \small
  \setlength{\tabcolsep}{5.0pt}
  \begin{tabular}{llrrrr}
\toprule
\rowcolor{TableHeader}
Model & Dataset & $N$ & ACC & \method & $\Delta$ \\
\midrule
Llama-3-8B & BoolQ & 3,270 & 76.54 & \bestcell{76.70} & \gaincell{+0.15} \\
 & MMLU & 14,042 & 65.00 & \bestcell{66.93} & \gaincell{+1.93} \\
 & BBH & 1,260 & 53.17 & \bestcell{56.67} & \gaincell{+3.49} \\
 & SciQ & 1,000 & 91.50 & \bestcell{93.40} & \gaincell{+1.90} \\
 & ARC & 3,548 & 89.04 & \bestcell{90.78} & \gaincell{+1.75} \\
\midrule
Qwen2.5-3B & BoolQ & 3,270 & 73.30 & \bestcell{77.71} & \gaincell{+4.40} \\
 & MMLU & 14,042 & 67.45 & \bestcell{68.37} & \gaincell{+0.92} \\
 & BBH & 1,260 & 51.98 & \bestcell{55.63} & \gaincell{+3.65} \\
 & SciQ & 1,000 & 91.80 & \bestcell{91.90} & \gaincell{+0.10} \\
 & ARC & 3,548 & 90.84 & \bestcell{92.42} & \gaincell{+1.58} \\
\midrule
Gemma-2-2B & BoolQ & 3,270 & \textbf{80.40} & \ourscell{80.24} & \losscell{-0.15} \\
 & MMLU & 14,042 & 58.62 & \bestcell{59.50} & \gaincell{+0.88} \\
 & BBH & 1,260 & 42.54 & \bestcell{44.05} & \gaincell{+1.51} \\
 & SciQ & 1,000 & 89.90 & \bestcell{91.80} & \gaincell{+1.90} \\
 & ARC & 3,548 & 85.82 & \bestcell{87.01} & \gaincell{+1.18} \\
\midrule
Phi-4-mini & BoolQ & 3,270 & 82.97 & \bestcell{85.26} & \gaincell{+2.29} \\
 & MMLU & 14,042 & 71.41 & \bestcell{73.04} & \gaincell{+1.62} \\
 & BBH & 1,260 & 57.46 & \bestcell{60.08} & \gaincell{+2.62} \\
 & SciQ & 1,000 & 91.80 & \bestcell{92.90} & \gaincell{+1.10} \\
 & ARC & 3,548 & 91.80 & \bestcell{93.21} & \gaincell{+1.41} \\
\midrule
Mistral-7B & BoolQ & 3,270 & 80.18 & \bestcell{83.88} & \gaincell{+3.70} \\
 & MMLU & 14,042 & 59.69 & \bestcell{62.09} & \gaincell{+2.40} \\
 & BBH & 1,260 & 46.35 & \bestcell{48.10} & \gaincell{+1.75} \\
 & SciQ & 1,000 & 88.60 & \bestcell{89.10} & \gaincell{+0.50} \\
 & ARC & 3,548 & 84.24 & \bestcell{87.01} & \gaincell{+2.76} \\
\bottomrule
\end{tabular}

  \caption{Complete matched comparison (accuracy, \%). The five rows within a
  model use different evaluation sets but the same trained policy family and
  decision rule.}
  \label{tab:main-deltas}
\end{table}

All 25 cells contain measured values; no missing cell is copied, interpolated,
or replaced by a macro average. The four transfer datasets contribute 20 of
the 25 comparisons, of which 19 improve over ACC. MMLU improves for every
backbone, while training uses only MMLU questions.

\section{Complete Component and Round Results}
\label{app:ablations}

The component comparisons follow recorded paths through the training pipeline. Pale
blue identifies the reported Full-R4 row, while black bold identifies
the strongest value in each comparable dataset column within a backbone.
All six completed configurations are reported. SFT+Trained-C combines the
role-SFT Actor with the trained Critic before Actor DPO. No-SFT starts Actors
and Critics from the base model and retains the three roles, preference
learning, private revision, and fixed vote. Full-R0 and Full-R4 use identical
trained policies and differ only in the evaluated Actor round.

The additional controls show why the training stages should be distinguished.
SFT+Trained-C reaches 77.79 on Qwen2.5 and 81.03 on Phi, exceeding Full-R4 by
0.58 and 0.14 macro points. No-SFT reaches 77.79 on Llama, exceeding Full-R4
by 0.89 points. Full-R4 is best on Gemma and Mistral and has the highest mean,
76.31, closely followed by SFT+Trained-C at 76.30. These descriptive results
support private revision while showing that the preferred initialization and
Actor update depend on the backbone.

The cell-level exceptions clarify the aggregate contrasts. Full-R4 improves on
Full-R0 in 23 cells; the only decreases are Qwen2.5 BBH ($-0.32$ points) and
Qwen2.5 SciQ ($-0.30$). Its largest revision gains occur on Phi BoolQ (+14.46),
Mistral BBH (+6.12), and Qwen2.5 BoolQ (+5.60). Adding an untrained base Critic
is similarly broad but not automatic: the SFT+Base-C contrast is negative on
Gemma BBH ($-1.03$) and Qwen2.5 SciQ ($-0.30$). Full-R4 improves on SFT-only in
24 cells, with Qwen2.5 SciQ ($-0.60$) as the sole exception.
Table~\ref{tab:full-ablations} includes both exceptions and all 24 positive
cells, and every reported mean uses the complete 25-cell matrix.

\begin{table}[H]
  \centering
  \small
  \setlength{\tabcolsep}{4.2pt}
  \begin{tabular}{llrrrrrr}
\toprule
\rowcolor{TableHeader}
Model & Variant & BoolQ & MMLU & BBH & SciQ & ARC & Macro \\
\midrule
Llama-3-8B & SFT-only & 73.82 & 60.58 & 49.84 & 88.20 & 87.29 & 71.95 \\
 & SFT+Base-C & 78.47 & 64.48 & 56.03 & 92.90 & 89.85 & 76.35 \\
 & SFT+Trained-C & 78.99 & 65.57 & 55.71 & 92.80 & 90.53 & 76.72 \\
 & Full-R0 & 71.47 & 62.15 & 51.35 & 88.80 & 87.57 & 72.27 \\
 & No-SFT & \textbf{79.94} & \textbf{67.05} & \textbf{57.62} & \textbf{93.80} & 90.53 & \textbf{77.79} \\
\rowcolor{HighlightBlue}
 & Full-R4 & 76.70 & 66.93 & 56.67 & 93.40 & \textbf{90.78} & 76.90 \\
\midrule
Qwen2.5-3B & SFT-only & 73.70 & 66.76 & 53.73 & 92.50 & 91.18 & 75.57 \\
 & SFT+Base-C & 76.39 & 66.93 & 56.43 & 92.20 & 91.57 & 76.71 \\
 & SFT+Trained-C & 77.28 & 68.20 & \textbf{57.46} & \textbf{94.00} & 92.00 & \textbf{77.79} \\
 & Full-R0 & 72.11 & 68.36 & 55.95 & 92.20 & 91.29 & 75.98 \\
 & No-SFT & 75.66 & 68.31 & 50.95 & 93.20 & 91.77 & 75.98 \\
\rowcolor{HighlightBlue}
 & Full-R4 & \textbf{77.71} & \textbf{68.37} & 55.63 & 91.90 & \textbf{92.42} & 77.21 \\
\midrule
Gemma-2-2B & SFT-only & 74.77 & 56.00 & 42.70 & 90.10 & 84.33 & 69.58 \\
 & SFT+Base-C & \textbf{80.95} & 58.55 & 41.67 & 91.60 & 87.03 & 71.96 \\
 & SFT+Trained-C & 80.40 & 58.82 & 43.02 & \textbf{92.10} & \textbf{87.20} & 72.31 \\
 & Full-R0 & 75.17 & 57.37 & 41.98 & 89.50 & 84.47 & 69.70 \\
 & No-SFT & 80.49 & \textbf{59.85} & 43.17 & 90.50 & 85.88 & 71.98 \\
\rowcolor{HighlightBlue}
 & Full-R4 & 80.24 & 59.50 & \textbf{44.05} & 91.80 & 87.01 & \textbf{72.52} \\
\midrule
Phi-4-mini & SFT-only & 69.79 & 68.40 & 53.25 & 90.10 & 90.33 & 74.37 \\
 & SFT+Base-C & 84.07 & 71.82 & 60.56 & 92.20 & 92.81 & 80.29 \\
 & SFT+Trained-C & 85.57 & 72.23 & \textbf{61.90} & 92.70 & 92.76 & \textbf{81.03} \\
 & Full-R0 & 70.80 & 68.73 & 56.59 & 90.40 & 90.56 & 75.41 \\
 & No-SFT & \textbf{85.75} & 72.26 & 58.89 & 92.50 & 92.64 & 80.41 \\
\rowcolor{HighlightBlue}
 & Full-R4 & 85.26 & \textbf{73.04} & 60.08 & \textbf{92.90} & \textbf{93.21} & 80.90 \\
\midrule
Mistral-7B & SFT-only & 78.75 & 57.01 & 44.76 & 86.70 & 83.79 & 70.20 \\
 & SFT+Base-C & 83.33 & 60.05 & 48.10 & \textbf{89.30} & 86.10 & 73.38 \\
 & SFT+Trained-C & 83.36 & 60.72 & \textbf{48.73} & 88.60 & 86.89 & 73.66 \\
 & Full-R0 & 79.69 & 57.11 & 41.98 & 86.30 & 82.89 & 69.60 \\
 & No-SFT & \textbf{84.62} & 61.19 & 47.94 & 88.20 & 86.30 & 73.65 \\
\rowcolor{HighlightBlue}
 & Full-R4 & 83.88 & \textbf{62.09} & 48.10 & 89.10 & \textbf{87.01} & \textbf{74.03} \\
\bottomrule
\end{tabular}

  \caption{Complete component matrix (accuracy, \%). Black bold marks the best
  value per dataset and backbone; Macro weights all five datasets equally.}
  \label{tab:full-ablations}
\end{table}

\begin{table}[H]
  \centering
  \small
  \begin{tabular}{lrrrrr}
\toprule
\rowcolor{TableHeader}
Model & R0 & R1 & R2 & R3 & R4 \\
\midrule
Llama-3-8B & 72.27 & 76.12 & 76.54 & \bestcell{77.04} & \ourscell{76.90} \\
Qwen2.5-3B & 75.98 & 76.97 & 77.25 & \bestcell{77.35} & \ourscell{77.21} \\
Gemma-2-2B & 69.70 & 72.43 & \bestcell{72.65} & 72.32 & \ourscell{72.52} \\
Phi-4-mini & 75.41 & 80.73 & \bestcell{80.93} & 80.78 & \ourscell{80.90} \\
Mistral-7B & 69.60 & 73.28 & 73.79 & 73.89 & \bestcell{74.03} \\
\midrule
\rowcolor{TableMacro}
Mean & 72.59 & 75.91 & 76.23 & 76.27 & \bestcell{76.31} \\
\bottomrule
\end{tabular}

  \caption{Round-wise macro accuracy (\%). R1 is the first Critic-conditioned
  revision. Pale blue marks the fixed endpoint; black bold marks the best round.}
  \label{tab:rounds}
\end{table}

Most of the five-round improvement arrives immediately. The mean gain is +3.32
points at R1, followed by +0.32 from R1 to R2, +0.04 from R2 to R3, and +0.04
from R3 to R4. The fixed R4 endpoint keeps one protocol across benchmarks, and
the recorded curve motivates future validation-set stopping rules.

\section{Role and Decision Diagnostics}
\label{app:diagnostics}

Table~\ref{tab:role-accuracies} reports the final accuracy of each role before
fusion. The best role varies across cells. Verification has the highest mean,
but Direct is strongest on Llama SciQ and ARC, Gemma BBH, Phi BBH and ARC,
and Mistral MMLU, BBH, and SciQ; Evidence is strongest on all five Qwen2.5
datasets, Phi MMLU, and Mistral BoolQ and ARC; and Direct and Evidence tie on
Phi SciQ. The fixed vote therefore
does not reduce to selecting one globally dominant role.

Agreement varies much more than parse reliability. Unanimity ranges from
40.16\% on Mistral BBH to 89.32\% on Phi-4-mini ARC, while two-of-three coverage
remains 87.46\% even in the lowest-coverage cell. Parse rate never falls below
99.76\%. The oracle-any-role gap ranges from 7.46 points for Phi to 10.83 for
Mistral, so headroom is not an artifact of one weak backbone or one parser
format.

Majority coverage is lowest on Qwen2.5 BBH (87.46\%) and highest on Phi
BoolQ (100.00\%). Fallback behavior is concentrated on the more heterogeneous
reasoning benchmark rather than uniformly distributed. The
 Evidence--Verification pair has the highest conditional accuracy (82.09\%)
 and agreement coverage (79.22\%), followed closely by Direct--Evidence at
79.17\% coverage. Evidence--Verification leads both pairwise summaries.

\begin{table}[H]
  \centering
  \scriptsize
  \setlength{\tabcolsep}{3.7pt}
  \begin{tabular}{llrrrrr}
\toprule
\rowcolor{TableHeader}
Model & Dataset & Direct & Evidence & Verification & \method & Oracle-any \\
\midrule
Llama-3-8B & BoolQ & 74.19 & 71.56 & 80.52 & \ourscell{76.70} & \oraclecell{88.26} \\
 & MMLU & 64.86 & 64.56 & 65.23 & \ourscell{66.93} & \oraclecell{80.98} \\
 & BBH & 53.81 & 52.06 & 56.19 & \ourscell{56.67} & \oraclecell{75.63} \\
 & SciQ & 92.80 & 91.80 & 91.20 & \ourscell{93.40} & \oraclecell{96.60} \\
 & ARC & 90.02 & 89.04 & 89.29 & \ourscell{90.78} & \oraclecell{95.29} \\
\midrule
Qwen2.5-3B & BoolQ & 68.96 & 79.51 & 77.49 & \ourscell{77.71} & \oraclecell{88.13} \\
 & MMLU & 65.05 & 67.28 & 66.14 & \ourscell{68.37} & \oraclecell{81.38} \\
 & BBH & 49.92 & 57.62 & 52.94 & \ourscell{55.63} & \oraclecell{75.16} \\
 & SciQ & 89.90 & 92.30 & 90.50 & \ourscell{91.90} & \oraclecell{96.50} \\
 & ARC & 90.28 & 90.95 & 89.04 & \ourscell{92.42} & \oraclecell{96.62} \\
\midrule
Gemma-2-2B & BoolQ & 76.45 & 78.69 & 79.39 & \ourscell{80.24} & \oraclecell{89.97} \\
 & MMLU & 57.97 & 57.81 & 58.57 & \ourscell{59.50} & \oraclecell{72.82} \\
 & BBH & 43.41 & 42.30 & 42.38 & \ourscell{44.05} & \oraclecell{60.32} \\
 & SciQ & 89.40 & 89.00 & 90.60 & \ourscell{91.80} & \oraclecell{96.30} \\
 & ARC & 84.78 & 84.61 & 85.79 & \ourscell{87.01} & \oraclecell{92.98} \\
\midrule
Phi-4-mini & BoolQ & 82.57 & 83.06 & 84.59 & \ourscell{85.26} & \oraclecell{91.56} \\
 & MMLU & 70.91 & 71.05 & 70.63 & \ourscell{73.04} & \oraclecell{83.24} \\
 & BBH & 59.21 & 56.67 & 57.94 & \ourscell{60.08} & \oraclecell{74.21} \\
 & SciQ & 92.20 & 92.20 & 90.50 & \ourscell{92.90} & \oraclecell{96.50} \\
 & ARC & 92.62 & 91.66 & 91.63 & \ourscell{93.21} & \oraclecell{96.28} \\
\midrule
Mistral-7B & BoolQ & 82.08 & 82.32 & 81.68 & \ourscell{83.88} & \oraclecell{91.41} \\
 & MMLU & 60.03 & 59.21 & 59.30 & \ourscell{62.09} & \oraclecell{76.78} \\
 & BBH & 47.38 & 44.76 & 44.68 & \ourscell{48.10} & \oraclecell{67.38} \\
 & SciQ & 88.00 & 85.80 & 86.90 & \ourscell{89.10} & \oraclecell{94.70} \\
 & ARC & 84.36 & 84.58 & 84.50 & \ourscell{87.01} & \oraclecell{94.05} \\
\bottomrule
\end{tabular}

  \caption{Per-role, voted, and oracle-any-role accuracy (\%). Pale blue marks
  the actual \method decision; mint marks diagnostic oracle headroom.}
  \label{tab:role-accuracies}
\end{table}

\begin{table}[H]
  \centering
  \small
  \begin{tabular}{lrr}
\toprule
\rowcolor{TableHeader}
Role pair & Agreement coverage & Accuracy when agreeing \\
\midrule
Direct + Evidence & 79.17 & 81.80 \\
Direct + Verification & 78.70 & 82.04 \\
Evidence + Verification & \bestcell{79.22} & \bestcell{82.09} \\
\bottomrule
\end{tabular}

  \caption{Pairwise final-answer agreement averaged over all 25 cells (\%).
  Conditional accuracy evaluates the shared answer only on agreeing examples.}
  \label{tab:pair-agreement}
\end{table}

Across all cells, parse rates average 99.93\%, so the agreement pattern is
driven by answer differences rather than systematic extraction failure.

\begin{table}[H]
  \centering
  \scriptsize
  \setlength{\tabcolsep}{3.8pt}
  \begin{tabular}{llrrrrr}
\toprule
\rowcolor{TableHeader}
Model & Dataset & Majority & Unanimous & Oracle-any & Fallback & Parsed \\
\midrule
Llama-3-8B & BoolQ & \ourscell{99.42} & 73.06 & \bestcell{88.26} & 0.58 & 99.79 \\
 & MMLU & \ourscell{93.51} & 58.48 & \bestcell{80.98} & 6.49 & 99.94 \\
 & BBH & \ourscell{89.52} & 43.97 & \bestcell{75.63} & 10.48 & 99.76 \\
 & SciQ & \ourscell{99.20} & 89.10 & \bestcell{96.60} & 0.80 & 100.00 \\
 & ARC & \ourscell{98.70} & 85.96 & \bestcell{95.29} & 1.30 & 100.00 \\
\midrule
Qwen2.5-3B & BoolQ & \ourscell{99.88} & 71.90 & \bestcell{88.13} & 0.12 & 100.00 \\
 & MMLU & \ourscell{94.52} & 60.52 & \bestcell{81.38} & 5.48 & 99.99 \\
 & BBH & \ourscell{87.46} & 43.10 & \bestcell{75.16} & 12.54 & 100.00 \\
 & SciQ & \ourscell{99.40} & 86.60 & \bestcell{96.50} & 0.60 & 100.00 \\
 & ARC & \ourscell{98.62} & 83.74 & \bestcell{96.62} & 1.38 & 100.00 \\
\midrule
Gemma-2-2B & BoolQ & \ourscell{99.91} & 74.31 & \bestcell{89.97} & 0.09 & 99.94 \\
 & MMLU & \ourscell{95.20} & 59.98 & \bestcell{72.82} & 4.80 & 99.81 \\
 & BBH & \ourscell{94.68} & 47.62 & \bestcell{60.32} & 5.32 & 99.84 \\
 & SciQ & \ourscell{98.40} & 84.60 & \bestcell{96.30} & 1.60 & 99.80 \\
 & ARC & \ourscell{98.22} & 80.61 & \bestcell{92.98} & 1.78 & 99.92 \\
\midrule
Phi-4-mini & BoolQ & \ourscell{100.00} & 81.83 & \bestcell{91.56} & 0.00 & 100.00 \\
 & MMLU & \ourscell{96.59} & 67.93 & \bestcell{83.24} & 3.41 & 99.92 \\
 & BBH & \ourscell{92.54} & 56.51 & \bestcell{74.21} & 7.46 & 100.00 \\
 & SciQ & \ourscell{99.50} & 88.50 & \bestcell{96.50} & 0.50 & 100.00 \\
 & ARC & \ourscell{99.35} & 89.32 & \bestcell{96.28} & 0.65 & 99.97 \\
\midrule
Mistral-7B & BoolQ & \ourscell{99.91} & 79.24 & \bestcell{91.41} & 0.09 & 99.94 \\
 & MMLU & \ourscell{92.29} & 53.26 & \bestcell{76.78} & 7.71 & 99.96 \\
 & BBH & \ourscell{88.33} & 40.16 & \bestcell{67.38} & 11.67 & 99.76 \\
 & SciQ & \ourscell{97.50} & 81.30 & \bestcell{94.70} & 2.50 & 99.90 \\
 & ARC & \ourscell{97.63} & 76.97 & \bestcell{94.05} & 2.37 & 99.94 \\
\bottomrule
\end{tabular}

  \caption{Complete final-round decision diagnostics (\%). Majority is the
  fraction with a valid two-of-three answer; Unanimous requires all three
  normalized answers to agree; Fallback is the complement of Majority.}
  \label{tab:decision-diagnostics}
\end{table}

\section{Answer Extraction and Evaluation Integrity}
\label{app:extraction}

All methods share a deterministic, dataset-aware answer extractor. It applies
the following normalization rules in order.

\begin{enumerate}
  \item remove model-specific thinking blocks without editing visible answer
  text;
  \item prioritize explicit final-answer markers near the end of the response;
  \item for multiple-choice tasks, extract an option letter by checking
  final-marker, tail-claim, \mbox{then weak-tail patterns;}
  \item for yes/no tasks, accept constrained final labels and label-plus-text
  forms such as ``A: Yes'' and ``B: No''; and
  \item select the last recognized match within the highest-priority pattern
  group, or return an invalid parse when no pattern matches.
\end{enumerate}

The test suite covers answer-marker precedence, case normalization, punctuation,
option-letter extraction, label-plus-text BoolQ forms, thinking-block removal,
truncation metadata, and anti-overmatching. Anti-overmatching examples include
ordinary mentions of ``yes'' or ``no'' inside a rationale that do not declare a
final answer. Every metric file records the version of the parser that
produced it.

During development, the label-plus-text forms exposed an overly narrow BoolQ
rule. The correction was applied to the shared extractor, after which every
affected cached raw generation was re-scored. Training and generation outputs
were not selectively repeated. Result aggregation rejects missing values,
inconsistent sample counts, and incomplete trial grids, and every table value
is checked against the released CSV files at the reported precision.

\end{document}